\pdfoutput=1

\documentclass[11pt]{article}

\usepackage[letterpaper,margin=1in]{geometry}

\usepackage[T1]{fontenc}
\usepackage[utf8]{inputenc}
\usepackage{amsmath,amssymb}
\usepackage{graphicx}
\usepackage{float}
\usepackage{subcaption}
\usepackage{multirow}
\usepackage{booktabs}                
\usepackage{enumitem}
\usepackage{siunitx}
\usepackage[most]{tcolorbox}         
\usepackage{listings}
\usepackage[title]{appendix}
\usepackage{algorithm}
\usepackage{algpseudocode}           
\usepackage[hidelinks]{hyperref}     
\usepackage[capitalise]{cleveref}
\usepackage{natbib}                  
\title{\bfseries AgentMap: Joint Equivalence and Subsumption Discovery for Ontology Matching}

\author{
  Yiping Song$^{1}$ \quad Jiaoyan Chen$^{1}$ \quad Renate A. Schmidt$^{1}$ \quad Hui Yang$^{1}$ \quad Wen Zhang$^{2}$\\[4pt]
  \small $^{1}$Department of Computer Science, The University of Manchester, UK\\
  \small \texttt{\{jiaoyan.chen, renate.schmidt, hui.yang-2\}@manchester.ac.uk},\\ \texttt{yiping.song@postgrad.manchester.ac.uk}\\[2pt]
  \small $^{2}$School of Software Technology, Zhejiang University, China\\
  \small \texttt{zhang.wen@zju.edu.cn}
}
\date{}

\begin{document}

\maketitle

\begin{abstract}
Ontology matching (OM) has traditionally been formulated as either equivalence discovery or subsumption matching. The existing OM systems identify only one type of semantic correspondence and cannot simultaneously discover equivalence and subsumption mappings. In this paper, we introduce \textbf{Hybrid Ontology Matching (HOM)}, a new OM task that unifies equivalence and subsumption discovery, and accordingly propose a Large Language Model (LLM)-based multi-agent OM framework \textbf{AgentMap} that is implemented by a series of interdependent semantic decisions.
Given a concept in the source ontology, AgentMap integrates semantic retrieval, hierarchical search, and collaborative multi-agent LLM reasoning to progressively explore the target ontology, identifying either the equivalent concept, if one exists, or the most fine-grained subsumer.
We further extend four OM datasets for a HOM benchmark and evaluate AgentMap under hybrid, equivalence-only, and subsumption-only settings. Experimental results show that AgentMap achieves promising performance on the hybrid setting, and at the same time outperforms equivalence matching and subsumption matching baselines on the equivalence-only and subsumption-only settings, respectively.
\end{abstract}

\noindent\textbf{Keywords:} Ontology Matching, Large Language Models, Subsumption Matching, Multi-Agent Systems

\section{Introduction}
Ontologies have been widely used for knowledge representation in many domains, such as SNOMED CT for healthcare and FoodOn for food and agriculture \cite{bos2006snomed,dooley2018foodon,staab2013handbook}.
For knowledge integration, reuse and interpretation, ontology matching (OM), which is to discover pairs of concepts across ontologies with a specific relationship like equivalence and subsumption, has been widely investigated \cite{otero2015ontology}. 
However, different ontologies may use different terminologies and adopt conceptualisations with different levels of granularity, and thus accurate OM is challenging, requiring semantic interpretation beyond lexical and structural matching. 

OM systems have evolved from traditional lexical matching and rule-based approaches, such as LogMap~\cite{jimenez2011logmap} and AgreementMakerLight (AML)~\cite{faria_agreementmakerlight_2013}, to encoder-based pre-trained language model (PLM)-based approaches, such as BERTMap~\cite{he_bertmap_2022}, and more recently to generative LLM-based frameworks, such as GenOM~\cite{song2026genom}. Most of the current OM systems focus on equivalence matching, although there are also some OM systems like BERTSub~\cite{chen_contextual_2023} that aim to discover concept pairs with subsumption relationships. 
More importantly, to the best of our knowledge, there is a shortage of OM systems that simultaneously discover equivalence and subsumption mappings.

To close this gap, we introduce \textbf{Hybrid Ontology Matching (HOM)}, a new OM task that requires jointly discovering equivalence and subsumption correspondences for each source concept (formalised in the Problem Statement).

Accordingly, we propose a multi-agent OM framework named AgentMap, which decomposes HOM into a series of interdependent semantic reasoning and decision-making steps. This design is inspired by LLM agent paradigms that interleave reasoning with action~\cite{yao2022react}, refine an initial decision using feedback from subsequent verification steps~\cite{madaan2023self}, and route control between specialised agents via handoffs~\cite{openai_handoff}. AgentMap combines embedding-based candidate retrieval, hierarchy-aware LLM reasoning, and lexical matching with conflict resolution, decomposing reasoning into specialised agents for equivalence screening, equivalence verification, and subsumption discovery.

We further extend four existing OM datasets involving ontologies of the food and biomedicine domains for a benchmark of the new HOM task. The main contributions of this work can be summarised as follows:
\begin{itemize}
    \item We introduce the new task of  Hybrid Ontology Matching (HOM), which requires simultaneously considering equivalence and subsumption mappings. It better reflects the requirements of real-world knowledge integration and reuse than conventional OM tasks.

    \item We develop the framework AgentMap that decomposes the complex HOM task into a set of sub-tasks orchestrated by an agentic workflow, effectively integrating established ontology matching techniques with LLM-based reasoning.

    \item 
    We evaluate AgentMap on four HOM datasets, demonstrating consistent improvements over alternative LLM-based strategies. We further compare AgentMap with both classic and recent OM systems under the conventional equivalence-only and subsumption-only settings, where it also achieves superior performance.
\end{itemize}

\section{Problem Statement}

Existing OM research is aimed at either equivalence matching or subsumption matching. 
In this work, we formulate a new task termed \textbf{Hybrid Ontology Matching (HOM)}, where the OM system is expected to identify both equivalence and subsumption mappings simultaneously.

In particular, given a source named concept \(c_s\) from a source ontology  \(\mathcal{O}_s\) and a target ontology \(\mathcal{O}_t\), HOM aims to identify a mapping represented as a triple $m=(c_s,c_t,r)$, where \(c_t\) is a named concept from \(\mathcal{O}_t\) and
\(
r \in \{\texttt{equivalence},\texttt{subsumption}\}
\)
is the semantic relation between \(c_s\) and \(c_t\).
The equivalence mapping is prioritised, as it represents more fine-grained semantic association. 
Namely, the system is expected to identify \(c_t\) as the equivalent concept of \(c_s\) in \(\mathcal{O}_t\) if it exists; otherwise, the system is expected to identify \(c_t\) as the most specific subsumer of \(c_s\) in \(\mathcal{O}_t\) (i.e., \(c_s \sqsubseteq c_t\) and there exists no named concept \( c_t'\neq c_t\) in \(\mathcal{O}_t\) such that \(c_s \sqsubseteq c_t' \sqsubseteq c_t \)).

\section{AgentMap}

\begin{figure*}[t]
\centering
\includegraphics[width=\textwidth]{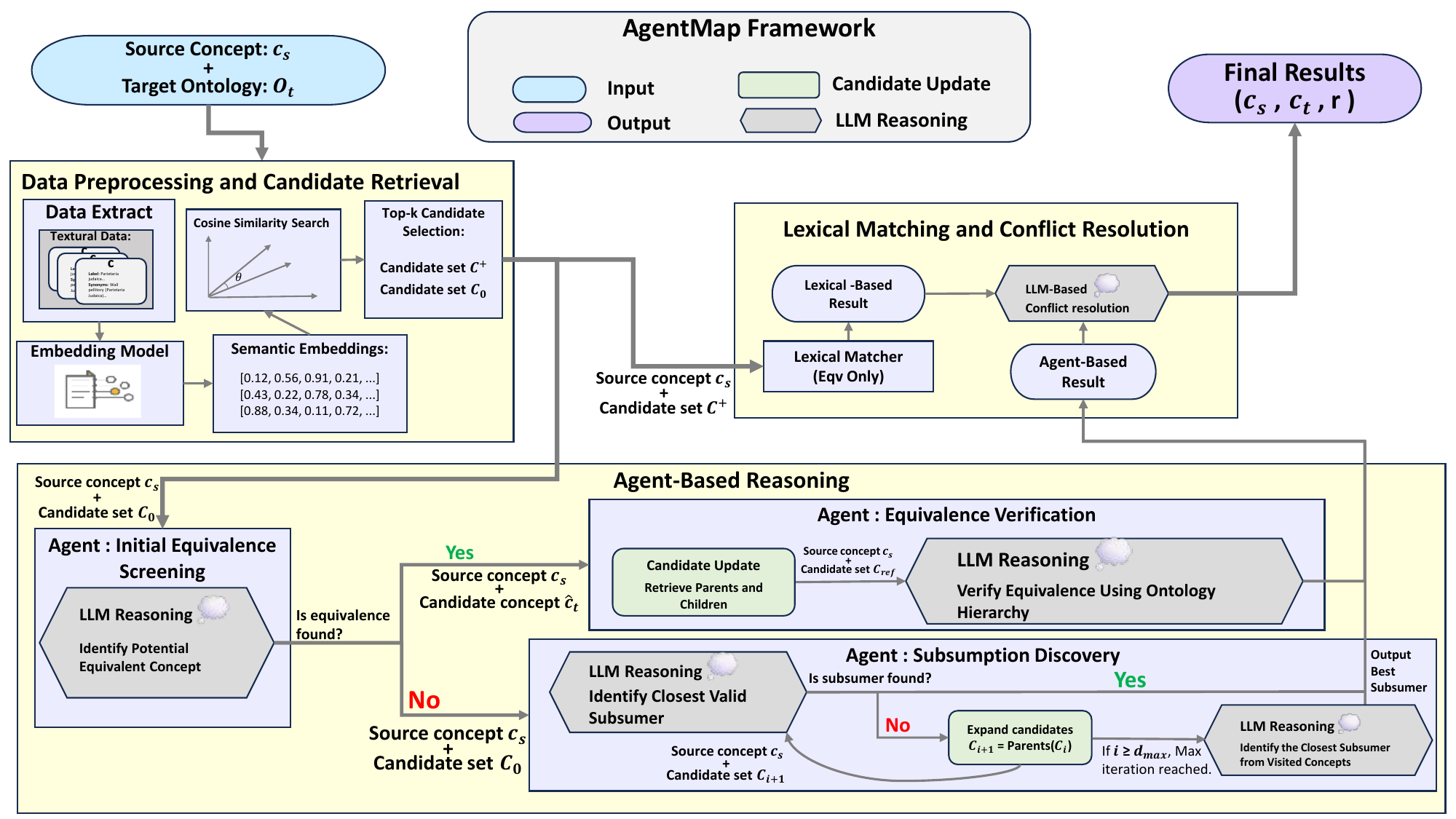}
\caption{Overview of the proposed AgentMap framework. Semantic retrieval first constructs task-specific candidate sets for the agent-based reasoning and lexical matching modules. Their predictions are subsequently reconciled through LLM-based conflict resolution to jointly predict the target concept and semantic relation.}
\label{fig:agentmap}
\end{figure*}

AgentMap addresses HOM by restricting LLM reasoning to a progressively refined candidate space, combining embedding-based retrieval with iterative exploration of the ontology hierarchy for efficient mapping discovery.
Figure~\ref{fig:agentmap} presents the overall architecture of AgentMap, which consists of three modules:

\textbf{Data Preprocessing and Candidate Retrieval} extracts textual information from the source concept and the target ontology, encodes concepts into semantic embeddings, performs cosine-similarity-based retrieval, and constructs two candidate sets for lexical matching and agent-based reasoning, respectively.

\textbf{Agent-Based Reasoning} performs collaborative reasoning through multiple LLM-based agents and ontology-aware search to progressively identify semantic correspondences between concepts.

\textbf{Lexical Matching and Conflict Resolution} complements agent-based semantic reasoning with lexical evidence and produces a unified correspondence by resolving potential inconsistencies between the two matching strategies.

\subsection{Data Preprocessing and Candidate Retrieval}

Given a source concept $c_s$ and a target ontology \(\mathcal{O}_t\), this module constructs two task-specific candidate sets for the agent and lexical matching modules.

AgentMap extracts the textual information of $c_s$ and the concepts in $\mathcal{O}_t$, specifically the label and synonyms of each concept, encodes them into semantic embeddings, and ranks the target concepts according to their cosine similarity to $c_s$. Two candidate sets are then constructed from this ranking using different top-k thresholds: (1) A smaller candidate set $C_0$ is constructed for the agent module, limiting costly LLM reasoning to the most semantically relevant concepts. (2) A larger candidate set $C^{+}$ is constructed for lexical matching, whose low computational cost permits broader candidate coverage. This dual candidate-set design provides each downstream module with a candidate scope suited to its computational characteristics and matching objective.

\subsection{Agent-Based Reasoning}

The agent-based reasoning module consists of three specialised agents: $Agent_{ES}$ for initial equivalence screening, $Agent_{EV}$ for equivalence verification, and $Agent_{SD}$ for subsumption discovery. The module follows an equivalence-first strategy. $Agent_{ES}$ first examines the initial candidate set $C_0$ for a potential equivalent concept. If an equivalence candidate is identified, $Agent_{EV}$ verifies the prediction through an ontology-structure-guided candidate update. Otherwise, $Agent_{SD}$ performs iterative subsumption discovery by progressively expanding the candidate set along the target ontology hierarchy.

\paragraph{Equivalence Discovery.}

$Agent_{ES}$ receives the source concept \(c_s\) and the initial candidate set \(C_0\), and performs LLM reasoning to determine whether \(C_0\) contains a valid equivalent target concept. If an equivalence candidate is identified, $Agent_{ES}$ outputs the most likely target concept \(\hat{c}_t\) and passes it to $Agent_{EV}$ for further verification.

$Agent_{EV}$ receives the source concept $c_s$ together with a candidate set $C_{\mathrm{ref}}$, constructed from the direct parent and child concepts of $\hat{c}_t$ in the target ontology. Specifically, AgentMap retrieves the direct parent and child concepts of $\hat{c}_t$, denoted $P(\hat{c}_t)$ and $Ch(\hat{c}_t)$, respectively, and, to avoid introducing every parent and child concept into the verification process, selects only the parent concept and the child concept that are most semantically similar to $c_s$:
\[
p^*
=
\arg\max_{p\in P(\hat{c}_t)}
sim(c_s,p),
\]
\[
ch^*
=
\arg\max_{ch\in Ch(\hat{c}_t)}
sim(c_s,ch),
\]
where $sim(\cdot,\cdot)$ denotes cosine similarity of the embeddings.

The resulting candidate set is
\[
C_{\mathrm{ref}}
=
\{\hat{c}_t,p^*,ch^*\}.
\]
If $\hat{c}_t$ has no direct parent concept (respectively, no direct child concept), $p^*$ (respectively, $ch^*$) is undefined and is simply omitted from $C_{\mathrm{ref}}$; in this case $C_{\mathrm{ref}}$ contains only $\hat{c}_t$ together with whichever of $p^*$ and $ch^*$ exists.

$Agent_{EV}$ then performs LLM reasoning over $C_{\mathrm{ref}}$ to verify the equivalence correspondence, comparing $\hat{c}_t$ against its most semantically relevant parent and child concepts. Based on this comparison, $Agent_{EV}$ outputs a single final candidate from $C_{\mathrm{ref}}$: it either retains $\hat{c}_t$, or replaces it with $p^*$ or $ch^*$, whichever candidate's granularity better matches $c_s$.

\paragraph{Subsumption Discovery.}

If $Agent_{ES}$ does not identify a valid equivalent concept, AgentMap invokes $Agent_{SD}$ to iteratively search for the closest valid subsumer. $Agent_{SD}$ initially receives the source concept \(c_s\) together with the initial candidate set \(C_0\). 
At \(i^{th}\) iteration, it performs LLM reasoning over the current candidate set \(C_i\) to determine whether any candidate is a valid subsumer of \(c_s\). If a valid subsumer \(c_t\) is identified, AgentMap returns
\(
(c_s,c_t,\texttt{subsumption}).
\)
Otherwise, the candidate set is updated using the structure of the target ontology. Specifically, all direct parents of each concept in \(C_i\) are collected to form the candidate set for the next iteration:
\[
C_{i+1}
=
Parents(C_i)
=
\bigcup_{c\in C_i} P(c).
\]
$Agent_{SD}$ then performs the same reasoning process over \(C_{i+1}\). Therefore, each iteration examines the direct parents of the complete candidate set from the preceding iteration, allowing the search to move upward through the ontology hierarchy level by level.

This ontology-structure-guided update continues until a valid subsumer is identified or the maximum iterations \(d_{\max}\) is reached. If no valid subsumer is found after \(d_{\max}\) upward expansions, $Agent_{SD}$ performs a final LLM reasoning step over all candidates visited throughout the search process:
\[
C_0 \cup C_1 \cup \cdots \cup C_{d_{\max}}.
\]
The agent then selects the most appropriate subsumer from these visited concepts.

\subsection{Lexical Matching and Conflict Resolution}

In parallel with the agent-based reasoning module, AgentMap performs lexical matching to identify equivalence correspondences supported by lexical evidence. The lexical matching module operates on the candidate set \(C^{+}\) generated during candidate retrieval. For each candidate concept in \(C^{+}\), the module compares its representations with the source concept, including labels and synonyms. A candidate is regarded as a lexical match if one of its labels or synonyms matches with some label or synonym of the source concept. Because lexical matching alone cannot determine hierarchical relations, this module predicts only \texttt{equivalence} correspondences.

In the end, the lexical matching result and the agent reasoning result are reconciled through LLM-based conflict resolution if they are not identical: (i) if no candidate in $C^{+}$ satisfies the lexical matching criterion above (i.e., the lexical matching module identifies no equivalence candidate), AgentMap returns the agent reasoning result; (ii) if lexical matching outputs an equivalent concept while the agents infer a subsumer, AgentMap returns the lexical matching result, following the equivalence-first problem setting; (iii) if lexical matching and agent reasoning output two different equivalent target concepts, AgentMap performs an additional LLM reasoning step to determine the final output.

\section{Evaluation}
\subsection{Benchmark and Metrics}
\paragraph{Benchmark Construction}

Existing benchmarks, such as Bio-ML used by Ontology Alignment Evaluation Initiative (OAEI) \cite{he_machine_2022}, provide equivalence and subsumption reference mappings independently. 
They can only evaluate OM systems for either equivalence or subsumption matching. We therefore construct a dedicated benchmark for the new task of HOM.

The proposed benchmark is built upon four equivalence OM datasets (tasks), including three from OAEI Bio-ML (SNOMED--FMA--Body, SNOMED--NCIT--Pharm, and NCIT--DOID--Disease) which match medical ontologies, and the HeLiS-FoodOn dataset \cite{chen_contextual_2023} which mathches a health lifestyle ontology with the food ontology. Since HOM requires jointly evaluating equivalence and subsumption matching against a single target ontology, we reorganize the source concepts of the original equivalence benchmark into two disjoint subsets: one for equivalence evaluation and one for subsumption evaluation. Subsumption ground truth, however, is not independently annotated; instead, following~\cite{he_machine_2022}, it is derived from the equivalence mappings by taking the direct parent of each equivalence target as the corresponding subsumer.

This construction creates an ambiguity: for concepts assigned to the subsumption subset, their true equivalence target is still present in the original target ontology, even though the intended ground truth is now its parent. Since the equivalence target is, by definition, a more specific and semantically closer match than its parent, any system able to identify it would report it instead of the coarser subsumption target used as ground truth, rendering subsumption evaluation ill-defined on the unmodified ontology. To resolve this, we remove, from the original target ontology, only the equivalence target concepts of the source concepts assigned to the subsumption subset, and reattach their child concepts to the corresponding parent to preserve the original taxonomy. The resulting ontology, denoted $\mathcal{O}_t$, is used consistently by all methods (AgentMap and baselines) throughout the paper, and jointly supports equivalence-only, subsumption-only, and HOM evaluation on a single, shared target ontology. 

Each resulting HOM dataset consists of a source ontology, a target ontology and a test set of source concepts \(T_{test}\), each of which is annotated by exactly one ground-truth target concept and the corresponding matching type (\texttt{equivalence} or \texttt{subsumption}).
The testing subset with equivalence (resp. subsumption) target concept is denoted as \(T_{eq}\) (resp. \(T_{sub}\)).
See Table~\ref{tab:benchmark_statistics} for more dataset statistics.

\begin{table}[t]
\centering
\small
\begin{tabular}{lrrr}
\toprule
Dataset & \#\(T_{eq}\) & \#\(T_{sub}\) & \#\(T_{test}\)\\
\midrule
SNOMED--FMA--Body     & 3099 & 3072 & 6171 \\
SNOMED--NCIT--Pharm    & 1120 & 3545 & 4665 \\
NCIT--DOID--Disease      & 2920 & 1694 & 4614 \\
HeLiS--FoodOn   & 174  & 190  & 364 \\
\bottomrule
\end{tabular}
\caption{Statistics of the constructed HOM benchmark.}
\label{tab:benchmark_statistics}
\end{table}

\paragraph{Evaluation Metrics}

Three evaluation settings are considered: HOM, equivalence-alone, and subsumption-alone.
The \textbf{HOM setting} uses the whole test set \(T_{test}\).
A source concept is regarded as correctly processed only if both the target concept and the matching type are correctly identified. The overall performance is measured by \textbf{Overall$_{\text{Acc.}}$} which is the ratio of the corrected predicted source concepts among all the source concepts \(T_{test}\).
To further analyse performance on different matching types, we additionally report the accuracies \textbf{Eqv$_{\text{Acc.}}$} and \textbf{Sub$_{\text{Acc.}}$} on the equivalence and subsumption subsets of the whole testing set, i.e., \(T_{eq}\) and \(T_{sub}\), respectively.

In the HOM setting, the system does not know the matching type, while in the \textbf{equivalence or subsumption-alone setting}, we let the system know the matching type in advance, so as to fairly comparing AgentMap with existing OM systems for either equivalence or subsumption matching.
A source concept is regarded as correctly processed if its ground-truth target concept is identified. The equivalence matching performance is measured by \(Accuracy_{eq}\), which is the ratio of correctly processed source concepts among \(T_{eq}\), and the subsumption matching performance is measured by \(Accuracy_{sub}\) which is the ratio of correctly processed source concepts among \(T_{sub}\).

\subsection{Baselines}

\paragraph{HOM Setting.}
We construct four LLM-based baselines by varying the prompting strategy and the available candidate set.
Two prompting strategies are considered: direct prompting and Chain-of-Thought (CoT) prompting~\cite{Wei2022ChainOT}. Two candidate configurations are evaluated. The first uses the same reasoning candidate set \(C_0\) as AgentMap, allowing the LLM to directly predict both the target concept and the semantic relation. The second expands \(C_0\) by including the parent concepts (up to two ontology levels) and direct child concepts of each retrieved candidate, forming an expanded neighbourhood candidate set that approximates the maximum ontology neighbourhood explored by AgentMap.

Combining the two prompting strategies with the two candidate configurations yields four baselines:

\begin{itemize}[leftmargin=*]
\item \textbf{LLM + \(C_0\)}: Direct prompting over the reasoning candidate set.
\item \textbf{LLM + \(C_0\) + CoT}: Chain-of-Thought prompting over the reasoning candidate set.
\item \textbf{LLM + Neighbour}: Direct prompting over the expanded neighbourhood candidate set.
\item \textbf{LLM + Neighbour + CoT}: Chain-of-Thought prompting over the expanded neighbourhood candidate set.
\end{itemize}

These baselines control both the prompting strategy and the available candidate coverage, helping isolate the contribution of AgentMap's staged multi-agent reasoning process.

\paragraph{Equivalence-alone Setting.}
For equivalence OM, we compare AgentMap with representative methods from three categories. Traditional OM systems, LogMap~\cite{jimenez2011logmap} and AML~\cite{faria_agreementmakerlight_2013}, take two complete ontologies as input and produce a single global alignment for the whole ontology pair rather than a per-concept prediction. To adapt them to $T_{eq}$, we look up each source concept in this global alignment and use its mapped target, if any, as the prediction. In contrast, pre-trained language model methods, represented by BERTMap~\cite{he_bertmap_2022}, and LLM methods, represented by GenOM~\cite{song2026genom}, natively predict a target concept for each source concept individually, and are therefore evaluated directly on \(T_{eq}\) following the same protocol as AgentMap.

\paragraph{Subsumption-alone Setting.}
For subsumption OM, we compare AgentMap with five representative methods that score each candidate subsumer in an embedding space and rank them accordingly: \textbf{(1)} General sentence embedding methods for semantic similarity, such as SBERT and OpenAI \texttt{text-embedding-3-small} Embedding; \textbf{(2)} Fine-tuned hirerachy embeddings methods, such as  HiT~\cite{3737916.3738385} and OnT~\cite{10.1007/978-3-032-09527-5_24}; and \textbf{(3)} Fine-tuned PLM-based classification model, such as BERTSub~\cite{chen_contextual_2023}, which is a representative subsumption  OM system. For all baselines, accuracy is computed as Hit@1, i.e., whether the top-scored candidate matches the ground truth.
Among these baselines, embedding-based methods encode each concept once and reuse the encoding for every comparison, and can therefore consider all named concepts as candidates; BERTSub instead reruns the model for every source--candidate pair, making exhaustive scoring over the full ontology infeasible, and therefore requires a predefined candidate list as input. To ensure a fair comparison, we provide BERTSub with a candidate set consisting of the initial retrieved candidates $C_0$ together with their parent concepts expanded up to two ontology levels, matching the maximum search range explored by $Agent_{SD}$ during subsumption discovery. This controls candidate coverage while allowing different ranking and reasoning strategies to be compared.

\subsection{Experimental Setup}

Unless otherwise specified, AgentMap uses GPT-4.1-mini as the backbone LLM throughout all experiments. All LLM-based methods are evaluated with a decoding temperature of 0. Ontology concepts are encoded using the OpenAI \texttt{text-embedding-3-small} embedding model, and cosine similarity is used for candidate retrieval. The agent-based reasoning module operates on the top-5 retrieved candidates (\(C_0\)), while the lexical matching module uses the top-20 retrieved candidates (\(C^{+}\)). During ontology-guided subsumption search, the maximum upward traversal \(d_{\max}\) is set to 2 (i.e., up to the grandparent level of the initially retrieved target candidates in the ontology hierarchy). The same configuration is adopted throughout all experiments unless explicitly stated otherwise.

To evaluate the robustness of AgentMap, we additionally replace GPT-4.1-mini with several representative open-source LLMs. \footnote{See Appendix for LLM prompt templates, SBERT embedding results, $C_0$ size sensitivity, and further details; code and data will be released.}

\subsection{Experimental Results}

\paragraph{HOM Results.}

\begin{table}[t]
\centering
\small
\setlength{\tabcolsep}{3pt}         
\renewcommand{\arraystretch}{1}
\begin{tabular}{llccc}
\toprule
\textbf{Task} & \textbf{System} & \textbf{Overall$_{\text{Acc.}}$} & \textbf{Eqv$_{\text{Acc.}}$} & \textbf{Sub$_{\text{Acc.}}$} \\
\midrule
\multirow{5}{*}{\shortstack[l]{SNOMED\\-FMA\\-Body}}
& LLM+\(C_0\) & 0.536 & 0.936 & 0.132 \\
& LLM+\(C_0\)+CoT & 0.546 & 0.940 & 0.149 \\
& LLM+Neighbour & 0.581 & 0.944 & 0.215 \\
& LLM+Neighbour+CoT & \underline{0.634} & \underline{0.949} & \underline{0.318} \\
\cmidrule{2-5}
& \textbf{AgentMap} & \textbf{0.657} & \textbf{0.957} & \textbf{0.354} \\
\midrule
\multirow{5}{*}{\shortstack[l]{SNOMED\\-NCIT\\-Pharm}}
& LLM+\(C_0\) & 0.255 & \textbf{0.981} & 0.026 \\
& LLM+\(C_0\)+CoT & 0.258 & 0.976 & 0.031 \\
& LLM+Neighbour & 0.374 & \underline{0.978} & 0.184 \\
& LLM+Neighbour+CoT & \underline{0.454} & 0.976 & \underline{0.289} \\
\cmidrule{2-5}
& \textbf{AgentMap} & \textbf{0.523} & \textbf{0.981} & \textbf{0.378} \\
\midrule
\multirow{5}{*}{\shortstack[l]{NCIT\\-DOID\\-Disease}}
& LLM+\(C_0\) & 0.728 & 0.940 & 0.362 \\
& LLM+\(C_0\)+CoT & 0.736 & 0.930 & 0.400 \\
& LLM+Neighbour & 0.760 & \textbf{0.952} & 0.430 \\
& LLM+Neighbour+CoT & \underline{0.772} & \underline{0.944} & \underline{0.476} \\
\cmidrule{2-5}
& \textbf{AgentMap} & \textbf{0.787} & \underline{0.944} & \textbf{0.513} \\
\midrule
\multirow{5}{*}{\shortstack[l]{HeLiS\\-FoodOn}}
& LLM+\(C_0\) & 0.387 & 0.713 & 0.089 \\
& LLM+\(C_0\)+CoT & 0.418 & 0.718 & 0.142 \\
& LLM+Neighbour & 0.434 & \textbf{0.793} & 0.105 \\
& LLM+Neighbour+CoT & \textbf{0.461} & \textbf{0.793} & \underline{0.158} \\
\cmidrule{2-5}
& \textbf{AgentMap} & \underline{0.452} & \underline{0.753} & \textbf{0.174} \\
\bottomrule
\end{tabular}
\caption{HOM Results. All the methods (AgentMap and the baselines) use GPT-4.1-mini as the backbone LLM. Bold indicates the best result and underline indicates the second-best in each column.}
\label{tab:main_results}
\end{table}

Table~\ref{tab:main_results} reports results for the HOM setting. 
AgentMap achieves the highest overall accuracy on the three biomedical benchmarks (SNOMED-FMA-Body, SNOMED-NCIT-Pharm, and NCIT-DOID-Disease), and is only slightly behind the best baseline method on HeLiS–FoodOn.

Specifically, AgentMap  achieved consistently  best subsumption accuracy across all four datasets. For example, on SNOMED-NCIT-Pharm, AgentMap's subsumption accuracy is up to 30.8\% higher than the best baseline (0.378 vs.\ 0.289).  
These results show that decomposing HOM into staged semantic decisions, combined with iterative ontology-guided search, is substantially more effective than single-step reasoning over a fixed candidate set.

For equivalence accuracy, the gap between AgentMap and the other baselines is much smaller. In some cases, such as HeLiS–FoodOn, LLM+Neighbourhood achieves higher equivalence accuracy than AgentMap. This is due to the fact that the baseline's expanded neighbourhood decides equivalence and subsumption together in a single LLM call, so the candidate set it sees when making the equivalence decision already includes the parent, grandparent, and child concepts needed for subsumption. AgentMap's $Agent_{ES}$, in contrast, judges equivalence using only $C_0$, which is a much smaller set. With only 174 equivalence instances in HeLiS--FoodOn, this gap in candidate coverage is further amplified, noticeably lowering AgentMap's equivalence accuracy relative to the baseline.

\paragraph{Equivalence-alone Results.}

\begin{table}[t]
\centering
\small
\setlength{\tabcolsep}{1pt}
\begin{minipage}[t]{0.48\columnwidth}
\centering
\textbf{SNOMED-FMA}

\vspace{1mm}

\begin{tabular}{lc}
\toprule
Method & Acc$_{\text{eq}}$\\
\midrule
LogMap & 0.470\\
AML & 0.906\\
BERTMap & 0.598\\
GenOM(Qwen32B) & 0.774\\

\midrule
\textbf{AgentMap(Qwen32B)} & \underline{0.949}\\
\textbf{AgentMap(GPT)} & \textbf{0.957}\\
\bottomrule
\end{tabular}
\end{minipage}
\hfill
\begin{minipage}[t]{0.48\columnwidth}
\centering
\textbf{SNOMED-NCIT}

\vspace{1mm}

\begin{tabular}{lc}
\toprule
Method & Acc$_{\text{eq}}$\\
\midrule
LogMap & 0.901\\
AML & 0.919\\
BERTMap & 0.809\\
GenOM(Qwen32B) & 0.888\\

\midrule
\textbf{AgentMap(Qwen32B)} & \underline{0.978}\\
\textbf{AgentMap(GPT)} & \textbf{0.981}\\
\bottomrule
\end{tabular}
\end{minipage}

\vspace{2mm}

\begin{minipage}[t]{0.48\columnwidth}
\centering
\textbf{NCIT-DOID}

\vspace{1mm}

\begin{tabular}{lc}
\toprule
Method & Acc$_{\text{eq}}$\\
\midrule
LogMap & 0.651 \\
AML & 0.780\\
BERTMap & 0.779\\
GenOM(Qwen32B) & 0.889\\

\midrule
\textbf{AgentMap(Qwen32B)} & \underline{0.942}\\
\textbf{AgentMap(GPT)} & \textbf{0.944}\\
\bottomrule
\end{tabular}
\end{minipage}
\hfill
\begin{minipage}[t]{0.48\columnwidth}
\centering
\textbf{HeLiS-FoodOn}

\vspace{1mm}

\begin{tabular}{lc}
\toprule
Method & Acc$_{\text{eq}}$\\
\midrule
LogMap & 0.626\\
AML & 0.701\\
BERTMap & 0.391\\
GenOM(Qwen32B) & 0.707\\

\midrule
\textbf{AgentMap(Qwen32B)} & \textbf{0.753}\\
\textbf{AgentMap(GPT)} & \textbf{0.753}\\
\bottomrule
\end{tabular}
\end{minipage}

\caption{Comparison with existing ontology matching system on equivalence alignment. Bold indicates the best result and underline the second-best in each column.}
\label{tab:eq_results}
\end{table}

Table~\ref{tab:eq_results} compares AgentMap with representative OM systems on equivalence-alone matching. As GenOM relies on next-token probabilities for candidate selection, it is only evaluated with open-source LLMs; we thus additionally report AgentMap under the same backbone (Qwen2.5-32B-Instruct~\cite{qwen25}) for a fair comparison, retaining GPT-4.1-mini as the default elsewhere.

AgentMap achieves the best performance on all four benchmarks under both backbones. LogMap fluctuates sharply (0.470 to 0.901): as an earlier-generation system, it relies heavily on external lexicons for matching, whose coverage varies across domains, unlike BERT- and LLM-based methods that draw on learned semantic representations instead. BERTMap is competitive on the three medical benchmarks but collapses to 0.391 on HeLiS--FoodOn: it is built on BioClinicalBERT, a BERT variant pre-trained specifically on clinical and biomedical text, whose vocabulary and representations are tailored to medical terminology and therefore transfer poorly to the food and lifestyle domain of HeLiS--FoodOn. GenOM stays comparatively stable across all four, consistent with LLMs carrying broader, less domain-specific knowledge. Switching AgentMap's backbone from Qwen2.5-32B to GPT-4.1-mini yields only modest further gains, indicating the improvement stems mainly from the reasoning framework rather than the backbone LLM.

\paragraph{Subsumption-alone Results.}
\begin{table}[t]
\centering
\small
\begin{minipage}[t]{0.48\columnwidth}
\centering
\textbf{SNOMED--FMA--Body}

\vspace{1mm}

\begin{tabular}{lc}
\toprule
Method & Acc$_{\text{sub}}$\\
\midrule
OnT & 0.054\\
HiT & 0.037\\
SBERT & 0.063\\
OpenAI & 0.076\\
BERTSub & \underline{0.191}\\
\midrule
\textbf{AgentMap} & \textbf{0.401
}\\
\bottomrule
\end{tabular}
\end{minipage}
\hfill
\begin{minipage}[t]{0.48\columnwidth}
\centering
\textbf{SNOMED--NCIT--Pharm}

\vspace{1mm}

\begin{tabular}{lc}
\toprule
Method & Acc$_{\text{sub}}$\\
\midrule
OnT & 0.012\\
HiT & 0.003\\
SBERT & 0.004\\
OpenAI & 0.009\\
BERTSub & \underline{0.046}\\
\midrule
\textbf{AgentMap} & \textbf{0.398}\\
\bottomrule
\end{tabular}
\end{minipage}

\vspace{2mm}

\begin{minipage}[t]{0.48\columnwidth}
\centering
\textbf{NCIT--DOID--Disease}

\vspace{1mm}

\begin{tabular}{lc}
\toprule
Method & Acc$_{\text{sub}}$\\
\midrule
OnT & 0.225\\
HiT & 0.261\\
SBERT & 0.263\\
OpenAI & 0.246\\
BERTSub & \underline{0.336}\\
\midrule
\textbf{AgentMap} & \textbf{0.564}\\
\bottomrule
\end{tabular}
\end{minipage}
\hfill
\begin{minipage}[t]{0.48\columnwidth}
\centering
\textbf{HeLiS--FoodOn}

\vspace{1mm}

\begin{tabular}{lc}
\toprule
Method & Acc$_{\text{sub}}$\\
\midrule
OnT & 0.190\\
HiT & 0.100\\
SBERT & 0.090\\
OpenAI & 0.074\\
BERTSub & \underline{0.269}\\
\midrule
\textbf{AgentMap} & \textbf{0.278}\\
\bottomrule
\end{tabular}
\end{minipage}
\caption{Comparison with existing subsumption matching methods. Bold indicates the best result and underline the second-best in each column.}
\label{tab:sub_results}
\end{table}

Table~\ref{tab:sub_results} presents results that compare AgentMap with the subsumption OM baselines. AgentMap achieves the highest accuracy on all four benchmarks, with gains most pronounced on the three biomedical benchmarks; it improves the accuracy over the best baseline BERTSub from 0.191 to 0.401 on SNOMED-FMA-Body, from 0.046 to 0.398 on SNOMED-NCIT-Pharm, and from 0.336 to 0.564 on NCIT-DOID-Disease. There is also a small gain (from 0.269 to 0.278) on HeLiS-FoodOn.Note that BERTSub is given the the initial retrieved candidates $C_0$ as AgentMap (see Baselines Section).

The gain is especially large on SNOMED-NCIT-Pharm, where the top-ranked candidates are often near-synonymous, making it hard for embedding- or BERT-based scoring to separate the correct subsumer from its closest competitors; AgentMap's LLM-based reasoning is better able to resolve such fine-grained distinctions. More broadly, these results show that purely ranking is insufficient for subsumption matching and it requires progressively exploring the hierarchy through iterative agent-based reasoning.

\subsection{Ablation Study}
Table~\ref{tab:ablation} reports ablation results on SNOMED-FMA-Body, the largest of the four benchmarks, examining two components: the hierarchical search mechanism and the Lexical Matching and Conflict Resolution (LM\&CR) module.

The hierarchical search appear to be the primary contributor to AgentMap's subsumption performance. Removing hierarchical search can reduces overall accuracy by 15.7\% and subsumption accuracy by more than half (58.2\%), while equivalence accuracy is unaffected, 
Removing LM\&CR, in contrast, causes only a modest drop in overall (1.4\%) and equivalence accuracy (1.7\%), with no effect on subsumption, indicating that LM\&CR mainly refines equivalence prediction.

\begin{table}[t]
\centering

\small
\setlength{\tabcolsep}{2pt}
\renewcommand{\arraystretch}{1.05}

\begin{tabular}{
@{}l
S[table-format=1.3]@{\,}l
S[table-format=1.3]@{\,}l
S[table-format=1.3]@{\,}l
@{}
}
\toprule
\textbf{Method}
& \multicolumn{2}{c}{\textbf{Overall$_{\text{Acc.}}$}}
& \multicolumn{2}{c}{\textbf{Eqv$_{\text{Acc.}}$}}
& \multicolumn{2}{c}{\textbf{Sub$_{\text{Acc.}}$}}
\\
\midrule

AgentMap
& 0.657 & {}
& 0.957 & {}
& 0.354 & {}
\\

w/o LM\&CR
& 0.648 & {\scriptsize(-1.4\%)}
& 0.941 & {\scriptsize(-1.7\%)}
& 0.354 & {\scriptsize(-0.0\%)}
\\

w/o Hierarchical Search
& 0.554 & {\scriptsize(-15.7\%)}
& 0.957 & {\scriptsize(-0.0\%)}
& 0.148 & {\scriptsize(-58.2\%)}
\\

\bottomrule
\end{tabular}
\caption{Ablation study of AgentMap on the SNOMED-FMA-Body benchmark 
(LM\&CR denotes the Lexical Matching and Conflict Resolution module).}
\label{tab:ablation}

\end{table}

\subsection{Effect of Backbone LLMs}

\begin{table}[t]
\centering
\small
\setlength{\tabcolsep}{1pt}         
\renewcommand{\arraystretch}{1}
\begin{tabular}{llccc}
\toprule
\textbf{Task} &
\textbf{LLM} &
\textbf{Overall$_{\text{Acc.}}$} &
\textbf{Eqv$_{\text{Acc.}}$} &
\textbf{Sub$_{\text{Acc.}}$} \\
\midrule

\multirow{5}{*}{\shortstack[l]{SNOMED--\\FMA-Body}}
& GPT-4.1-mini          & \textbf{0.657} &  0.957& \textbf{0.354} \\
& Qwen2.5-32B-Instruct  & 0.635 & 0.949 & 0.318 \\
& Qwen2.5-72B-Instruct  & 0.639&  0.955&   0.320 \\
& Llama3.1-70B-Instruct & 0.641 & \textbf{0.960} & 0.318 \\
& Mixtral-8x7B-Instruct & 0.572 & 0.936 & 0.205   \\

\midrule

\multirow{5}{*}{\shortstack[l]{SNOMED--\\NCIT-Pharm}}
& GPT-4.1-mini          &  0.523 &  \textbf{0.981}&  0.378 \\
& Qwen2.5-32B-Instruct  & \textbf{0.528} &  0.978& \textbf{0.386}\\
& Qwen2.5-72B-Instruct  & 0.520 &   0.979 &0.375\\
& Llama3.1-70B-Instruct &   0.509   & 0.979 &  0.361    \\
& Mixtral-8x7B-Instruct &  0.459 &  0.973  &  0.296  \\

\midrule

\multirow{5}{*}{\shortstack[l]{NCIT--DOID\\-Disease}}
& GPT-4.1-mini          & \textbf{0.787} & \textbf{0.944} & \textbf{0.513}  \\
& Qwen2.5-32B-Instruct  &  0.772& 0.942& 0.478\\
& Qwen2.5-72B-Instruct  &  0.777 & 0.941  & 0.493  \\
& Llama3.1-70B-Instruct &   0.755&  0.940 &  0.434    \\
& Mixtral-8x7B-Instruct &   0.693&  0.914 &   0.311   \\

\midrule

\multirow{5}{*}{\shortstack[l]{HeLiS--\\FoodOn}}
& GPT-4.1-mini          & 0.452 & \textbf{0.753} &  0.174 \\
& Qwen2.5-32B-Instruct  &  0.434&  \textbf{0.753}& 0.142 \\
& Qwen2.5-72B-Instruct  &   \textbf{0.453}&  \textbf{0.753} &   \textbf{0.179}\\
& Llama3.1-70B-Instruct &   0.405 &    0.730&   0.106   \\
& Mixtral-8x7B-Instruct &    0.371&  0.695 &  0.074    \\

\bottomrule
\end{tabular}
\caption{Effect of different backbone LLMs on HOM, Bold indicates the best result.}
\label{tab:llm_analysis}
\end{table}

Table~\ref{tab:llm_analysis} evaluates the effect of backbone LLMs on AgentMap. The framework performs stably across LLMs such as GPT-4.1-mini, Qwen2.5-32B-Instruct, Qwen2.5-72B-Instruct~\cite{qwen25}, and Llama3.1-70B-Instruct~\cite{dubey2024llama}, indicating that its effectiveness is largely independent of the backbone. GPT-4.1-mini and the Qwen2.5 variants are the strongest overall, each best on two of the four benchmarks, with Llama3.1-70B-Instruct remaining competitive; 
Superisingly, in SNOMED-NCIT-Pharm task, Qwen2.5-32B even archives better performance than bigger Qwen2.5-72B model. 
Mixtral-8x7B-Instruct~\cite{jiang2024mixtral}, in contrast, consistently underperforms across all four, which may be due to its weaker capabilities.

\section{Related Work}

\subsection{Equivalence Ontology Matching}

Equivalence OM has evolved from rule-based systems to neural representation learning and, more recently, LLM-based reasoning. Early systems, such as LogMap~\cite{jimenez2011logmap} and AML~\cite{faria_agreementmakerlight_2013}, combine lexical similarity, ontology structures, and logical reasoning to construct high-quality ontology alignments, and remain strong baselines in the OAEI benchmark.

With the development of deep learning, OM has increasingly relied on learned semantic representations~\cite{he_bertmap_2022}. Early neural approaches employed convolutional neural networks (CNNs) to encode ontology concepts from textual descriptions~\cite{bento-etal-2020-ontology}. More recent methods adopt transformer-based language models, including BERTMap~\cite{he_bertmap_2022}, BioSTransformer~\cite{menad_biostransformers_2023}, BioGITOM~\cite{oulefki_biogitom_2025}, and Magneto~\cite{liu_magneto_2025}, which leverage contextual language representations, graph neural architectures, or hybrid small-large language models to improve semantic matching accuracy.

Recent advances in LLMs have further shifted OM from representation learning toward semantic reasoning~\cite{he_exploring_2023}. Representative systems, including GenOM~\cite{song2026genom}, LogMap-LLM~\cite{lushnei-etal-2026-large}, Olala~\cite{hertling_olala_2023}, and LLM4OM~\cite{giglou_llms4om_2024}, exploit the reasoning capability of LLMs to perform ontology alignment through semantic understanding and multi-step inference.

Although these methods differ substantially in architecture, they are all designed for equivalence OM, where the objective is to identify concepts referring to the same real-world entity across heterogeneous ontologies.

\subsection{Subsumption Ontology Matching}

Compared with equivalence OM, subsumption OM has received considerably less attention. Existing studies commonly formulate the task as a candidate ranking problem~\cite{he_machine_2022}, where the objective is to identify the correct subsumer from a predefined candidate list.

BERTSub~\cite{chen_contextual_2023} is one of the few approaches specifically designed for subsumption OM. It uses contextual language models to rank candidate subsumers under existing benchmark settings. However, BERTSub assumes that the benchmark-supported subsumer is already contained in the candidate list and therefore evaluates candidate ranking rather than ontology-wide subsumption discovery.

Ontology representation learning methods such as OnT~\cite{10.1007/978-3-032-09527-5_24} and HiT~\cite{3737916.3738385} have also been used as concept encoders in related ranking settings. However, they are not designed specifically for subsumption OM. In this work, we include them only as encoder-based baselines to examine how ontology-aware representations perform under our evaluation protocol.

HOM instead requires searching the target ontology to jointly identify the target concept and determine whether the relation is \texttt{equivalence} or \texttt{subsumption} — a joint formulation that, to the best of our knowledge, no existing OM framework supports.

\section{Conclusion}

We introduced Hybrid Ontology Matching (HOM), reframing equivalence and subsumption discovery as a single task evaluated jointly, reflecting the fact that, in practice, whether a source concept has an exact match or only a broader one is not known in advance. AgentMap addresses HOM by decomposing this joint decision into staged, interdependent agent reasoning steps rather than a single LLM judgment. Our ablation and cross-backbone experiments show that this staged decomposition, rather than candidate coverage or backbone strength, drives AgentMap's gains, suggesting iterative, structure-aware reasoning as a general principle for LLM agents over hierarchical structures. AgentMap also sets a new state of the art on subsumption matching, though its absolute accuracy remains below 0.5 on three of four benchmarks, showing this sub-task is still considerably harder than equivalence matching. Future work includes closing this gap through more targeted hierarchical search and extending HOM to richer semantic relations.

\bibliography{AgentMap_arxiv}

\appendix

\section{Implementation Details}
The benchmark is split into equivalence and subsumption evaluation subsets using a fixed random seed of 42 to ensure reproducibility. All open-source models, including the open-source backbone LLMs (Qwen2.5-32B-Instruct, Qwen2.5-72B-Instruct, Llama3.1-70B-Instruct, Mixtral-8x7B-Instruct) and the local embedding, equivalence and subsumption baselines are run on two NVIDIA A100 GPUs.

\section{Prompt Templates}

\subsection{Agent Initial Equivalence Screening Prompt}
\begin{tcolorbox}[
    colback=gray!5,
    colframe=black!60,
    title=\textbf{Agent A Prompt: Initial Equivalence Screening (System Message)},
    fonttitle=\bfseries\small,
    breakable,
    boxrule=0.5pt,
    left=4pt, right=4pt, top=4pt, bottom=4pt
]
\small
Given one source concept and a list of candidate concepts.

\vspace{2pt}
\textbf{Task:}\\
Determine whether any candidate is the correct equivalent concept of the source. Equivalent means:
\begin{itemize}[leftmargin=*, itemsep=0pt, topsep=2pt]
    \item refers to the same real-world entity or concept
    \item same meaning
    \item same level of specificity
\end{itemize}
If no exact equivalent exists, select \texttt{NONE}.

\vspace{2pt}
\textbf{Rules:}
\begin{itemize}[leftmargin=*, itemsep=0pt, topsep=2pt]
    \item Select exactly one candidate ID, or \texttt{NONE}.
    \item Do not invent candidate IDs.
\end{itemize}

\vspace{2pt}
Return exactly this format:
\begin{verbatim}
Reasoning:
<short reasoning>
Selected:
<candidate_id or NONE>
\end{verbatim}

\vspace{4pt}
\textit{\footnotesize The source concept and candidate list are supplied via the user message at inference time.}
\end{tcolorbox}

\subsection{Agent Equivalence Verification Prompt}
\begin{tcolorbox}[
    colback=gray!5,
    colframe=black!60,
    title=\textbf{Agent Prompt: Equivalence Verification (System Message)},
    fonttitle=\bfseries\small,
    breakable,
    boxrule=0.5pt,
    left=4pt, right=4pt, top=4pt, bottom=4pt
]
\small
Given one source concept and a small list of candidate concepts.

\vspace{2pt}
\textbf{Task:}\\
Select the single best candidate as the final equivalence result. The correct result should:
\begin{itemize}[leftmargin=*, itemsep=0pt, topsep=2pt]
    \item have the same core meaning as the source
    \item match the source granularity best
\end{itemize}

\vspace{2pt}
\textbf{Rules:}
\begin{itemize}[leftmargin=*, itemsep=0pt, topsep=2pt]
    \item Always select exactly one candidate ID from the given list.
    \item Do not invent candidate IDs.
\end{itemize}

\vspace{2pt}
Return exactly this format and no extra text:
\begin{verbatim}
Reasoning:
<short reasoning>
Selected:
<candidate_id>
\end{verbatim}

\vspace{4pt}
\textit{\footnotesize The source concept and the updated candidate set $C_{\text{ref}} = \{\hat{c}_t, p^*, ch^*\}$ are supplied via the user message at inference time.}
\end{tcolorbox}

\subsection{Agent Subsumption Discovery Prompt}
\begin{tcolorbox}[
    colback=gray!5,
    colframe=black!60,
    title=\textbf{Agent Prompt (Iterative Search): Subsumption Discovery (System Message)},
    fonttitle=\bfseries\small,
    breakable,
    boxrule=0.5pt,
    left=4pt, right=4pt, top=4pt, bottom=4pt
]
\small
Given one source concept and a list of candidate concepts.

\vspace{2pt}
\textbf{Task:}\\
Choose the nearest broader candidate for the source. If none is clearly broader, choose \texttt{NONE}.

\vspace{2pt}
\textbf{Rules:}
\begin{itemize}[leftmargin=*, itemsep=0pt, topsep=2pt]
    \item A valid choice must be broader than the source.
    \item It must be the closest available parent-level concept.
    \item Do not choose candidates that are only related, part-based, sibling-level, or more specific.
    \item Use only the given labels and synonyms.
    \item Choose exactly one candidate ID, or \texttt{NONE}.
\end{itemize}

\vspace{2pt}
Return exactly this format:
\begin{verbatim}
Reasoning:
<short reasoning>
Selected:
<candidate_id or NONE>
\end{verbatim}

\vspace{4pt}
\textit{\footnotesize The source concept and the current candidate set $C_i$ are supplied via the user message at inference time. If \texttt{NONE} is returned, the candidate set is expanded to $C_{i+1} = Parents(C_i)$ and this prompt is reapplied.}
\end{tcolorbox}

\vspace{6pt}

\begin{tcolorbox}[
    colback=gray!5,
    colframe=black!60,
    title=\textbf{Agent Prompt (Fallback): Final Subsumer Selection (System Message)},
    fonttitle=\bfseries\small,
    breakable,
    boxrule=0.5pt,
    left=4pt, right=4pt, top=4pt, bottom=4pt
]
\small
Given one source concept and a list of candidate concepts.

\vspace{2pt}
\textbf{Task:}\\
Select the candidate that is the closest broader concept of the source concept.

\vspace{2pt}
\textbf{Definitions:}
\begin{itemize}[leftmargin=*, itemsep=0pt, topsep=2pt]
    \item A broader concept is a parent-level or ancestor-level concept that can subsume the source.
    \item The selected candidate should be semantically broader than the source, not equivalent to it, not narrower than it, and not merely related to it.
    \item If multiple candidates are broader, select the most specific and closest broader candidate.
    \item You must select exactly one candidate from the list.
\end{itemize}

\vspace{2pt}
\textbf{Reasoning procedure:}
\begin{enumerate}[leftmargin=*, itemsep=0pt, topsep=2pt]
    \item Identify whether each plausible candidate is broader than the source.
    \item Exclude candidates that are equivalent, narrower, sibling concepts, parts, attributes, or merely related concepts.
    \item Among the remaining broader candidates, choose the closest and most specific one.
    \item If no candidate is a perfect direct parent, choose the best available broader candidate.
\end{enumerate}

\vspace{2pt}
Return exactly this format:
\begin{verbatim}
Reasoning:
<short reasoning>
Selected:
<candidate_id>
\end{verbatim}

\vspace{4pt}
\textit{\footnotesize This prompt is applied once, after $d_{\max}$ upward expansions, over the complete search trajectory $C_0 \cup C_1 \cup \cdots \cup C_{d_{\max}}$.}
\end{tcolorbox}

\subsection{Conflict Resolution Prompt: Equivalence Arbitration}
\begin{tcolorbox}[
    colback=gray!5,
    colframe=black!60,
    title=\textbf{Conflict Resolution Prompt: Equivalence Arbitration (System Message)},
    fonttitle=\bfseries\small,
    breakable,
    boxrule=0.5pt,
    left=4pt, right=4pt, top=4pt, bottom=4pt
]
\small
You are an ontology alignment arbitration agent. Your task is to choose which target concept is more appropriate as the equivalence match for the source concept. Choose exactly one option: \texttt{A} or \texttt{B}.

\vspace{2pt}
\textbf{Definitions:}
\begin{itemize}[leftmargin=*, itemsep=0pt, topsep=2pt]
    \item \texttt{equivalence} means that the source concept and target concept refer to the same or nearly identical concept at the same conceptual granularity.
\end{itemize}

\vspace{2pt}
\textbf{Rules:}
\begin{itemize}[leftmargin=*, itemsep=0pt, topsep=2pt]
    \item Use only labels and synonyms.
    \item Do not use or infer from IRIs.
    \item Do not introduce a new target concept.
    \item Do not output subsumption.
    \item Briefly explain the conflict.
    \item Then output the final choice.
\end{itemize}

\vspace{2pt}
Return exactly this format:
\begin{verbatim}
Reason:
<short reasoning>
Choice:
A or B
\end{verbatim}

\vspace{4pt}
\textit{\footnotesize This prompt is invoked when the lexical matching module and the agent-based reasoning module output two different equivalence target concepts (options A and B, supplied via the user message), following the third case of the conflict resolution rules in the Lexical Matching and Conflict Resolution section.}
\end{tcolorbox}

\subsection{HOM Baseline Prompts}

\begin{tcolorbox}[
    colback=gray!5,
    colframe=black!60,
    title=\textbf{Direct Prompting Baseline (System Message)},
    fonttitle=\bfseries\small,
    breakable,
    boxrule=0.5pt,
    left=4pt, right=4pt, top=4pt, bottom=4pt
]
\small
Given one source concept and a list of target candidate concepts.

\vspace{2pt}
\textbf{Task:}\\
Select exactly one candidate and assign exactly one relation: \texttt{equivalence} or \texttt{subsumption}.

\vspace{2pt}
\textbf{Definitions:}
\begin{itemize}[leftmargin=*, itemsep=0pt, topsep=2pt]
    \item \texttt{equivalence} means the candidate refers to the same real world concept as the source.
    \item \texttt{subsumption} means the candidate is broader than the source and can act as a direct or near direct parent level concept.
\end{itemize}

\vspace{2pt}
\textbf{Rules:}
\begin{itemize}[leftmargin=*, itemsep=0pt, topsep=2pt]
    \item Choose \texttt{equivalence} if one candidate has the same meaning as the source.
    \item If no equivalent candidate exists, choose the nearest broader candidate.
    \item Do not choose siblings, children, parts, or merely related concepts.
    \item Prefer the most specific broader candidate when choosing \texttt{subsumption}.
    \item You must choose exactly one candidate from the given list.
    \item Use only the provided labels and synonyms.
    \item Candidate IDs are only placeholders. They do not contain semantic information.
\end{itemize}

\vspace{2pt}
Output JSON only:
\begin{verbatim}
{
"candidate_id": "...",
"relation": "equivalence" or "subsumption"
}
\end{verbatim}

\vspace{4pt}
\textit{\footnotesize The source concept and the candidate set are supplied via the user message at inference time. This prompt is used for both the LLM + $C_0$ and LLM + Neighbourhood baselines, differing only in the candidate set provided.}
\end{tcolorbox}

\vspace{6pt}

\begin{tcolorbox}[
    colback=gray!5,
    colframe=black!60,
    title=\textbf{Chain-of-Thought Prompting Baseline (System Message)},
    fonttitle=\bfseries\small,
    breakable,
    boxrule=0.5pt,
    left=4pt, right=4pt, top=4pt, bottom=4pt
]
\small
Given one source concept and a list of target candidate concepts.

\vspace{2pt}
\textbf{Task:}\\
Select exactly one candidate and assign exactly one relation: \texttt{equivalence} or \texttt{subsumption}.

\vspace{2pt}
\textbf{Definitions:}
\begin{itemize}[leftmargin=*, itemsep=0pt, topsep=2pt]
    \item \texttt{equivalence} means the candidate refers to the same real world concept as the source.
    \item \texttt{subsumption} means the candidate is broader than the source and can act as a direct or near direct parent level concept.
\end{itemize}

\vspace{2pt}
\textbf{Rules:}
\begin{itemize}[leftmargin=*, itemsep=0pt, topsep=2pt]
    \item Choose \texttt{equivalence} if one candidate has the same meaning as the source.
    \item If no equivalent candidate exists, choose the nearest broader candidate.
    \item Do not choose siblings, children, parts, or merely related concepts.
    \item Prefer the most specific broader candidate when choosing \texttt{subsumption}.
    \item You must choose exactly one candidate from the given list.
    \item Use only the provided labels and synonyms.
    \item Candidate IDs are only placeholders. They do not contain semantic information.
    \item Keep the reasoning concise.
    \item The reasoning should be natural language.
    \item The final answer must be a JSON object.
\end{itemize}

\vspace{2pt}
Output format:
\begin{verbatim}
Reasoning:
<concise natural-language reasoning>

Final answer:
{
"candidate_id": "...",
"relation": "equivalence" or "subsumption"
}
\end{verbatim}

\vspace{4pt}
\textit{\footnotesize The source concept and the candidate set are supplied via the user message at inference time. This prompt is used for both the LLM + $C_0$ + CoT and LLM + Neighbourhood + CoT baselines, differing only in the candidate set provided.}
\end{tcolorbox}

\section{Effect of Open-Source Resources}
\subsection{Closed-Source vs.\ Open-Source Configurations}
Table~\ref{tab:closed_open} reports results on SNOMED--FMA--Body under four configurations, combining OpenAI or SBERT embeddings with GPT-4.1-mini or Qwen2.5-32B-Instruct as the backbone. The fully closed-source configuration achieves the best performance (0.657 overall), while the fully open-source configuration (SBERT + Qwen2.5-32B-Instruct) trails by 13.3 points (0.524), a moderate gap given the removal of all commercial components. Comparing the two hybrid configurations suggests that the embedding model contributes more to this gap than the backbone LLM: replacing the backbone alone (OpenAI + Qwen2.5-32B-Instruct) costs only 2.2 points relative to the closed-source setting, whereas replacing the embedding model alone (SBERT + GPT-4.1-mini) costs 11.2 points, driven primarily by a large drop in equivalence accuracy (0.957 to 0.777). We examine this effect of embedding choice in more detail across all four benchmarks in the next subsection.

\begin{table*}[t]
\centering
\caption{Performance of AgentMap under closed-source, hybrid, and open-source settings on the SNOMED--FMA Body task.}
\label{tab:closed_open}
\renewcommand{\arraystretch}{1.7}
\begin{tabular}{lccc}
\hline
\textbf{Configuration} & \textbf{Overall$_{\text{Acc.}}$} & \textbf{Eqv$_{\text{Acc.}}$} & \textbf{Sub$_{\text{Acc.}}$} \\
\hline
{ Closed-source (OpenAI Emb. + GPT-4.1-mini)}
    & \textbf{0.657} & \textbf{0.957} & \textbf{0.354} \\

{ Hybrid (OpenAI Emb. + Qwen2.5-32B-Instruct)}
    & 0.635 & 0.949 & 0.318 \\

{ Hybrid (SBERT + GPT-4.1-mini)}
    & 0.545 & 0.777 & 0.312 \\

{ Open-source (SBERT + Qwen2.5-32B-Instruct)}
    & 0.524 & 0.750 & 0.296 \\
\hline
\end{tabular}
\end{table*}

\subsection{Effect of Embedding Model Across Benchmarks}
While Table~\ref{tab:closed_open} controls for both factors on a single benchmark, we further isolate the effect of the embedding model alone by fixing the backbone to Qwen2.5-32B-Instruct across all four benchmarks. Table~\ref{tab:embedding_comparison} reports the results.

OpenAI embeddings outperform SBERT on both equivalence and subsumption accuracy on three of the four benchmarks (Body, Pharm, and Disease), indicating that retrieval quality generally affects both stages of AgentMap. However, which stage is more affected varies substantially across datasets: on SNOMED--FMA--Body, the embedding choice affects equivalence more (a 21.0\% relative drop with SBERT) than subsumption (6.9 \%), whereas on SNOMED--NCIT--Pharm the pattern reverses sharply, with subsumption accuracy dropping by over half (52.3\%) while equivalence remains largely unaffected (0.6\%). NCIT--DOID--Disease shows a more balanced effect on both metrics. This inconsistency suggests that the relative sensitivity of each stage to retrieval quality is dataset-dependent, though identifying the specific underlying factors is beyond the scope of this analysis.

On HeLiS--FoodOn, the trend reverses entirely: SBERT achieves higher overall and subsumption accuracy than OpenAI embeddings, while OpenAI retains an advantage on equivalence. Given the small size of this dataset (174 equivalence and 190 subsumption instances), this reversal likely reflects sampling variance rather than a systematic advantage of SBERT. Overall, these results indicate that while AgentMap remains functional with a fully open-source embedding model, retrieval quality has a measurable and dataset-dependent impact on both equivalence and subsumption performance.
\begin{table*}[t]
\centering
\caption{Effect of embedding model (OpenAI \texttt{text-embedding-3-small} vs.\ SBERT \texttt{all-MiniLM-L6-v2}) on HOM performance across the four benchmarks, with Qwen2.5-32B-Instruct fixed as the backbone LLM. Bold indicates the better result within each dataset.}
\label{tab:embedding_comparison}
\small
\setlength{\tabcolsep}{4pt}
\begin{tabular}{llccc}
\toprule
\textbf{Dataset} & \textbf{Embedding} & \textbf{Overall} & \textbf{Eqv} & \textbf{Sub} \\
\midrule
\multirow{2}{*}{SNOMED--FMA--Body}
& OpenAI & \textbf{0.635} & \textbf{0.949} & \textbf{0.318} \\
& SBERT  & 0.524 & 0.750 & 0.296 \\
\midrule
\multirow{2}{*}{SNOMED--NCIT--Pharm}
& OpenAI & \textbf{0.528} & \textbf{0.978} & \textbf{0.386} \\
& SBERT  & 0.374 & 0.972 & 0.184 \\
\midrule
\multirow{2}{*}{NCIT--DOID--Disease}
& OpenAI & \textbf{0.772} & \textbf{0.942} & \textbf{0.478} \\
& SBERT  & 0.735 & 0.920 & 0.416 \\
\midrule
\multirow{2}{*}{HeLiS--FoodOn}
& OpenAI & 0.434 & \textbf{0.753} & 0.142 \\
& SBERT  & \textbf{0.442} & 0.701 & \textbf{0.205} \\
\bottomrule
\end{tabular}
\end{table*}

\section{Effect of Candidate Set Size (Top-$k$)}
\begin{figure*}[t]
\centering

\begin{minipage}[t]{0.48\textwidth}
\centering
\includegraphics[width=\linewidth]{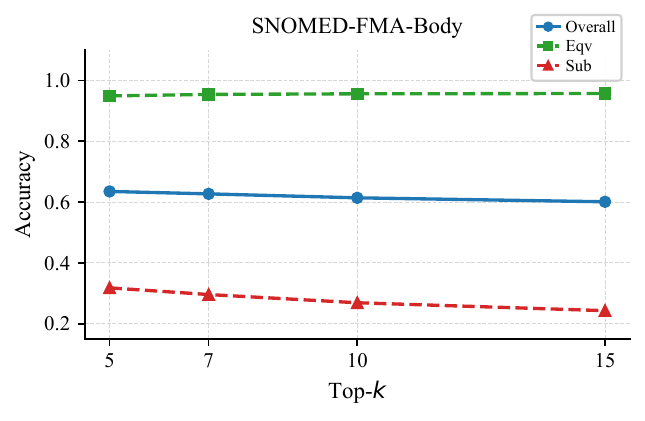}
\small (a) SNOMED--FMA--Body
\end{minipage}
\hfill
\begin{minipage}[t]{0.48\textwidth}
\centering
\includegraphics[width=\linewidth]{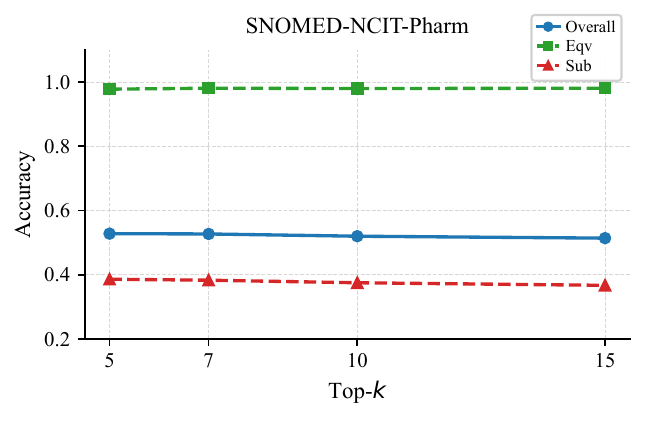}
\small (b) SNOMED--NCIT--Pharm
\end{minipage}

\vspace{4mm}

\begin{minipage}[t]{0.48\textwidth}
\centering
\includegraphics[width=\linewidth]{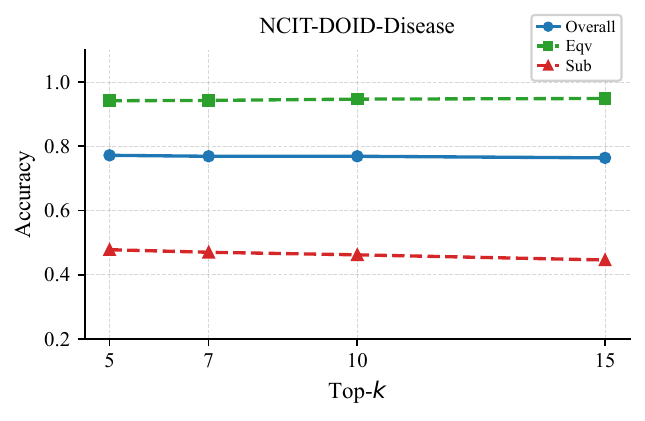}
\small (c) NCIT--DOID--Disease
\end{minipage}
\hfill
\begin{minipage}[t]{0.48\textwidth}
\centering
\includegraphics[width=\linewidth]{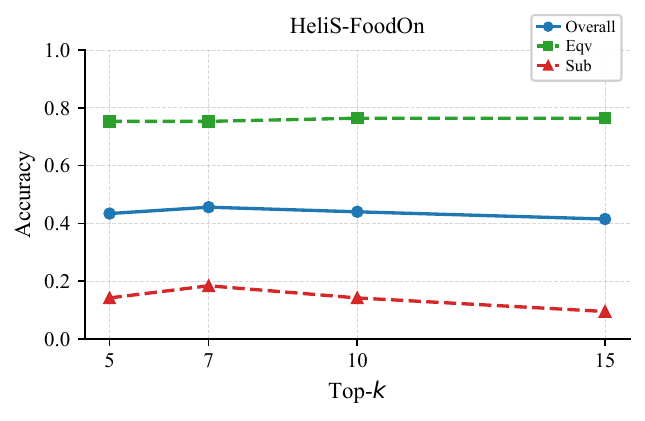}
\small (d) HeLiS--FoodOn
\end{minipage}

\caption{Effect of candidate set size $k$ on HOM results across the four benchmarks, using Qwen2.5-32B-Instruct as the backbone.}
\label{fig:topk_all}
\end{figure*}

We further examine the sensitivity of AgentMap to the size of the agent reasoning candidate set $C_0$, using Qwen2.5-32B-Instruct as the backbone, with $k \in \{5, 7, 10, 15\}$. Figure~\ref{fig:topk_all} reports the results across all four benchmarks.

As $k$ increases, Overall and Sub accuracy decrease monotonically, while Eqv accuracy rises only marginally. This indicates that a larger candidate set does not meaningfully improve equivalence discovery, since the correct equivalent concept is typically already covered by a small $k$, but it introduces more distractor candidates into Agent C's subsumption search, increasing the likelihood of selecting an incorrect broader concept. This trend is consistent across the three biomedical benchmarks; HeLiS--FoodOn shows a similar pattern but with more fluctuation, reflecting its much smaller test set size. These results support our choice of $k=5$ as the default configuration, which achieves the best overall and subsumption accuracy while minimising the LLM reasoning cost associated with a larger candidate set.

\end{document}